\documentclass{article}

\usepackage{arxiv-cns-2021}

\usepackage[utf8]{inputenc}

\title{Hippocampal Spatial Mapping As Fast Graph Learning}
\author{
  Marcus Lewis \\
  Numenta \\
  \texttt{mrcslws@gmail.com}
}

\usepackage{natbib}
\usepackage{graphicx}

\usepackage{hyperref}
\hypersetup{
    colorlinks=true,
    linkcolor=black,
    citecolor=black,
    filecolor=black,
    urlcolor=black,
}

\begin{document}

\maketitle

\begin{abstract}
The hippocampal formation is thought to learn spatial maps of environments, and in many models this learning process consists of forming a sensory association for each location in the environment. This is inefficient, akin to learning a large lookup table for each environment. Spatial maps can be learned much more efficiently if the maps instead consist of arrangements of sparse environment parts. In this work, I approach spatial mapping as a problem of learning graphs of environment parts. Each node in the learned graph, represented by hippocampal engram cells, is associated with feature information in lateral entorhinal cortex (LEC) and location information in medial entorhinal cortex (MEC) using empirically observed neuron types. Each edge in the graph represents the relation between two parts, and it is associated with coarse displacement information. This core idea of associating arbitrary information with nodes and edges is not inherently spatial, so this proposed fast-relation-graph-learning algorithm can expand to incorporate many spatial and non-spatial tasks.
\end{abstract}


\section{Introduction}

The hippocampal formation has been shown to be involved in spatial mapping \citep{OKeefe1971, Taube1990, Hafting2005}. Analogies to geographic maps or computer graphics suggest that spatial mapping should involve associating a Euclidean space of unique location representations, for example Cartesian coordinates, with a set of locations in the real world. In neuroscience, such a Euclidean space is often theorized to be provided by entorhinal grid cells \citep{Hafting2005}. A single grid cell module can represent Euclidean space ambiguously, while an array of grid cells can theoretically create unique location representations \citep{Fiete2008}. In models that simulate spatial mapping of the brain at a low level, the population activity of multiple grid cell modules is often used in this way \citep{Lewis2019, Whittington2020}, or they use a single grid cell module to complement traditional Cartesian coordinates \citep{Milford2008}. These models imply that the brain maps space in a way that is fundamentally similar to metrical geographic maps or computer graphics. On the other hand, in cognitive-inspired robotics \citep{Kuipers2000}, spatial mapping is less about creating a single coherent Cartesian map, and more about representing relationships between environment segments, with a patchwork of small Cartesian maps optionally learned on top of this system.

In this paper I argue that the multi-module grid cell code is actually not ideal as an analog of long-range Cartesian coordinates for spatial mapping. Instead, I suggest that the hippocampal formation approaches spatial mapping as graph-learning, quickly learning graphs which store relations between environment parts, and I suggest that grid cells provide additional spatial information that is not captured by these relations. This alternate approach to spatial mapping loses some generalization benefits that come naturally with Cartesian coordinates, but this approach also has more potential for flexibility. I argue that fast graph learning, with a learned repertoire of relations, is a more likely fundamental algorithm for the hippocampal formation than Cartesian-like mapping.

\section{Model}

In this section I present the theoretical and empirical motivations for graph-based spatial mapping, then I describe the general model at a high level. Finally, I present a possible low-level implementation built mainly from empirically observed neuron types.

\subsection{Problems with using only grid cells for spatial mapping}

There are theoretical and empirical reasons to doubt that the brain uses the multi-module grid cell code to create an analog of Cartesian coordinates.

\begin{figure}[]
\begin{center}
\includegraphics[width=0.9\textwidth]{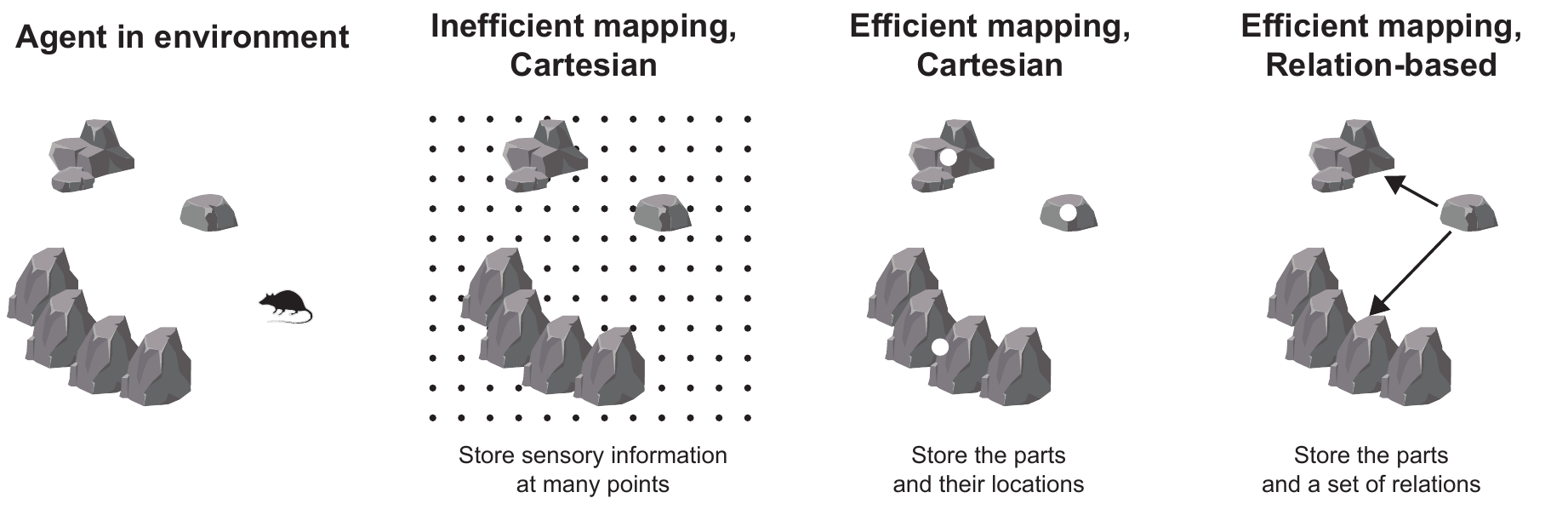}
\end{center}
\caption{\label{fig:efficient-mapping}\textbf{Efficient mapping.} An agent (a rodent) observes an environment and learns a map. One mapping strategy is to visit each location in the environment and record what you sense. This is inefficient, as indicated by the number of dots in the second panel. It is much more efficient to split the environment into parts or chunks and somehow store their relative locations. One strategy, common in computer graphics, is to store Cartesian coordinates for each part. Another strategy is to instead store a set of parts and a set of relations between them.}
\end{figure}

Theoretically, the multi-module grid cell code is initially appealing if you approach mapping as a process of moving from location to location and associating sensory input with those locations \citep{Lewis2019, Whittington2020}. However, this approach to mapping is inefficient in that it requires visiting every location in the environment. A more practical and less expensive mapping algorithm would represent environments as arrangements of parts, using sensory input to quickly infer that arrangement of parts \textbf{(Figure \ref{fig:efficient-mapping})}. This requires the agent to be able to detect locations of these ``parts'' from a distance. Thus, it becomes important for grid cells to support the two computational operations of adding a displacement to a location and inferring the displacement between two locations. It is controversial whether grid cells support inferring or applying spatial displacements; some show that it is theoretically possible \citep{Bush2015}, while others argue that it is too difficult \citep{Fiete2008}, requiring an impractically large lookup table. Without such an operation, grid cells lose much of their desirability as a foundation for a map, especially given the fact that this also eliminates the ability to infer large navigation vectors. In previous work \citep{Hawkins2019}, my collaborators and I attempted to solve efficient mapping with the multi-module grid cell code by posing spatial mapping as composition of grid cell maps, associating one map with each environment ``part'' and quickly detecting the relationships of these maps. The two downsides of this approach were that it could only translate the maps relative to each other, not rotate or scale them, and that it had no notion of distance between the agent and the sensed ``part'', eliminating opportunities for incorporating this distance into sensory processing (for example, scaling the image on the agent's retina according to the distance).

Empirically, there are multiple reasons to doubt that grid cell modules are working closely together to create unique location codes. They are anatomically separated \citep{Stensola2012}, and anatomical studies have not yet found any significant population of neurons receiving projections from more than one or two modules. In electrophysiological recordings, grid cell modules appear to scale and distort independent of one another \citep{Stensola2012}. The anatomical separation issue has been addressed by models by relying instead on horizontal connections in the hippocampus \citep{Whittington2020}, but the independent scaling and distortion would not be expected in a population location code. In particular, independent distortion of grids eliminates even the theoretical possibility of detecting spatial displacements between locations.

Thus, grid cells likely play a role in spatial mapping, but there are reasons to doubt that the brain uses multiple grid cell modules to create a Cartesian-like map.

\subsection{Using relations for spatial mapping}

\begin{figure}[]
\begin{center}
\includegraphics[width=0.9\textwidth]{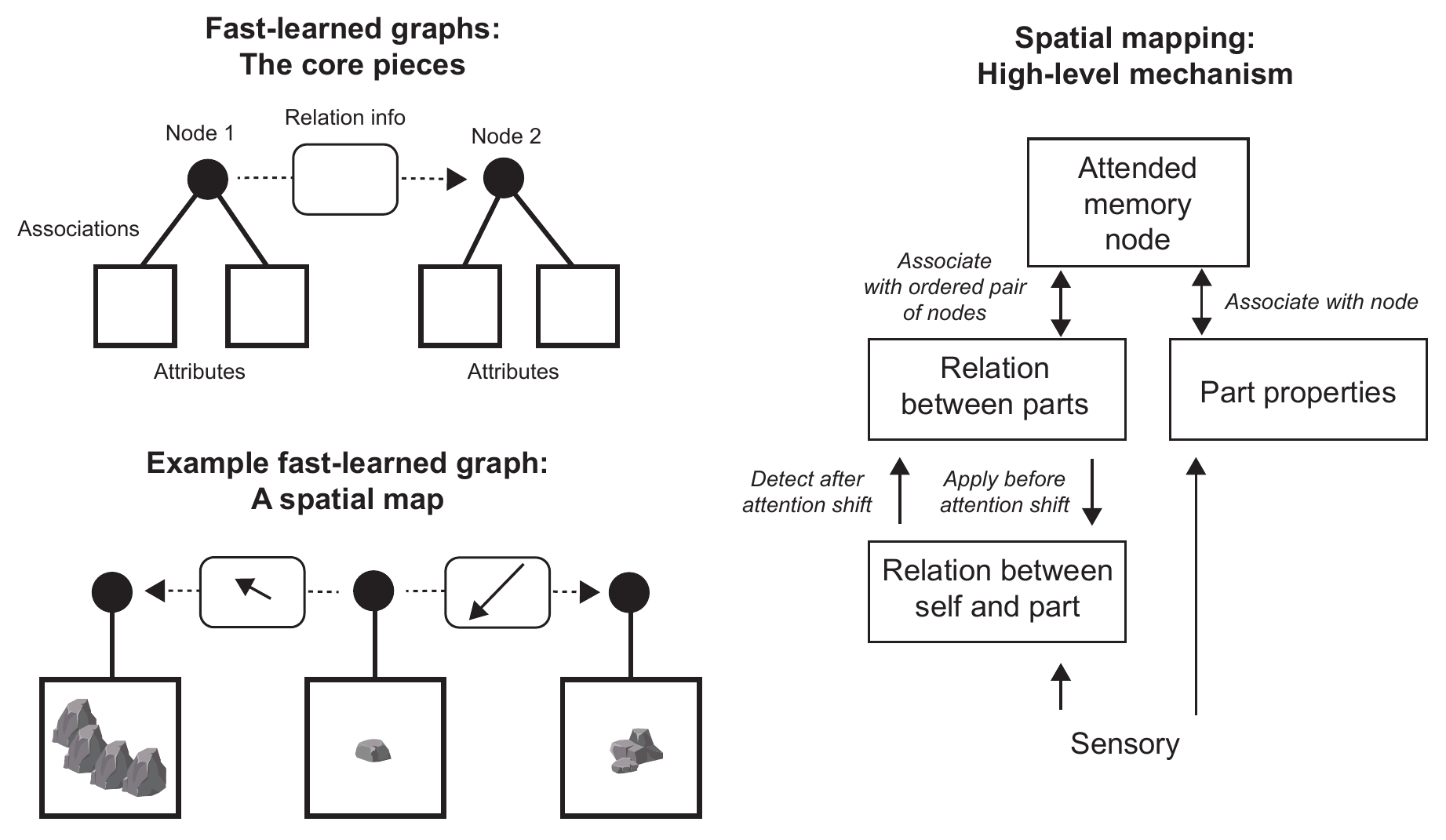}
\end{center}
\caption{\label{fig:graphs}\textbf{Fast-learned graphs as a general data structure.} The system stores in memory a set of nodes, associations with those nodes, a set of edges, and associations with those edges. This figure shows a graph-based map of the environment from Figure \ref{fig:efficient-mapping}. A neural system could build graph-based spatial maps by attending to different parts, assigning nodes to each part, and detecting relations between attended parts and associating those relations with edges. After a graph has been learned, it can be used for prediction (via the down arrows).}
\end{figure}

Instead of associating parts in the environment with unique Cartesian-like locations, the hippocampal formation could represent environments as a set of relations between parts. In this view, an environment is not represented as ``\texttt{Part 1 at (x1, y1), Part 2 at (x2, y2), Part 3 at (x3, y3), ...}'', but instead as ``\texttt{Part 1 related to Part 2 with (relation1), Part 3 related to Part 2 with (relation2), ...}''. This representation is fundamentally a graph \textbf{(Figure \ref{fig:graphs})}. Processing input by building up graph-like data structures was an early idea in the history of AI \citep{Minsky1974} and is highly general. ``Scene graphs'' are an active area of research in computer vision \citep{Chang2021}.

During spatial mapping, these graphs could be quickly built up from a few glances. \textbf{Figure \ref{fig:graphs}} shows a high-level mechanism. The agent uses its sensory input to infer a description of the sensed environment part as well as the spatial relationship between the agent and the part. A graph node is allocated to this part, and it is associated with the part's description. The agent attends to another environment part, activating a different description and spatial relationship, and allocating a second graph node. From this pair of agent-part spatial relationships, the agent infers the spatial relationship between the two parts, and this relation is associated with this ordered pair of graph nodes. Now that this graph has been learned, it can be used to guide attention shifts and to predict what will be sensed after each attention shift. The size of the map scales with the number of environment parts. In environments with many parts, the number of parts can be reduced by chunking them into aggregate parts.

The upside of this approach is that it is flexible and can easily apply to both spatial and non-spatial domains. The set of relations between nodes can be fully learned; in spatial domains, they could simply be spatial displacement vectors, while in other domains they could be more similar to the relations that are expressed in language.

The downside of this approach, relative to the Cartesian map, is that it does not provide easy direct access to relations of arbitrary pairs of parts. When parts are assigned Cartesian locations, the displacement between any two parts can be trivially computed, but if instead a graph of relations between parts is learned, detecting the relation between a pair of parts requires the relation to have been previously learned, or it must be inferred by finding a path between the parts on the graph and aggregating the relations. The ``Tolman-Eichenbaum Machine'' \citep{Whittington2020} aims to get the best of both of these worlds, applying to non-spatial domains while working similarly to Cartesian coordinates, essentially associating information with individual graph nodes. In this paper, I instead propose that the hippocampal formation associates information with directed edges between nodes, and this admittedly makes such generalization require more advanced computation over graphs. This is further discussed in section \ref{section:graph-processing}.

In the domain of spatial mapping, the set of relations will tend to leave some information ambiguous. When the relations are simple displacement vectors, those displacement vectors will tend to be coarse for two reasons. First, these graphs should be built up very quickly, and given the uncertainty inherent to sensory input this is much easier if the displacements are coarse. Second, using neurons to represent and decode fine-grained displacement vectors would require a lot of neural tissue, but with coarse displacements a small number of neurons and synapses could represent displacements over medium-to-large distances. The ambiguity of these relations is where grid cells will often become useful. Grid cells themselves are ambiguous, but in a different way. They can be viewed as filling in the information that was is left ambiguous by the relation graph. This principle is elaborated in in section \ref{section:combined-building-blocks}.

Thus in this theory, the hippocampus quickly learns graphs, associating information with nodes and edges of the graph. Relations are associated with edges. Grid cell locations, among other things, are associated with nodes providing extra information that was not provided by the relation. Alternately, we can invert this narrative. Grid cells solve one part of representing location, but they do it ambiguously and in a way that does not provide good support for representing displacements, and those aspects are supported by fast-learned relation graphs.

\subsection{Detailed example low-level model}

Here I describe a neural implementation of a system that learns environments as relation graphs, mainly using cell types empirically observed in the hippocampal formation. This section walks through the building blocks of this model then describes how they are combined. The model is human-interpretable; once the model is understood, the reader will be able to mentally simulate it.

Multiple design decisions of this model were informed by empirical evidence. For example, how should the spatial relationship between the self and the sensed part be represented? Another example: should the conversion from egocentric to allocentric locations occur before detecting relations between parts, or after? From a theoretical standpoint these questions do not have single obviously-correct answers. Based on the empirical existence of object-vector cells (described below), this model uses object-vector cells and performs the coordinate transformation before detecting relations. The actual cortex may use a hybrid approach, and the approach may vary by species.

\subsubsection{A short review of existing neural building blocks}

This model is built up from a set of components that have each been empirically observed and/or used previously in other models. These components are reviewed below.

\textbf{Head-direction cells:} These cells fire when an animal is facing a particular direction relative to the environment \citep{Taube1990}. They are analogous to the compass rose of a map. Head direction cells are mapped onto environment directions using sensory heuristics, for example detecting the main axis of the environment and anchoring relative to that axis \citep{Julian2018b}.

\textbf{Object-vector cells:} These cells fire when an animal is at a particular distance and environment-centric direction from any object \citep{Hoydal2019}. They can be thought of as a reusable set of location representations that can be bound to any object, but they use head direction cells as their compass rose rather than anchoring to the orientation of the object. Object-vector cells and grid cells both rely on the brain performing a coordinate transformation based on head direction cells. An animal's sensory and movement information are naturally encoded in a viewer-centric or egocentric reference frame, and head-direction cells are needed to convert these sensory vectors and movement vectors into an environment-centric or allocentric reference frame \citep{Bicanski2018}. This coordinate transformation in this model largely mirrors the model of \cite{Bicanski2018}. Head direction cells provide a routing signal, specifying how vectors must be rotated to convert between viewer-centric and environment-centric directions. Object-vector cells are just one of many types of ``-vector'' cells observed in the hippocampal formation \citep{Bicanski2020}. For simplicity, this model only focuses on object-vector cells, but many other vector cells could be substituted.

\textbf{Grid cells:} These cells fire when an animal is at particular locations in the environment. The set of firing locations forms a triangular or hexagonal ``grid''. The firing patterns of different grid cells in a local area of neural tissue are similar, shifted / translated versions of the same grid. A cluster of grid cells with a complete set of translations of the same grid are known as a ``grid cell module''. This model uses a single grid cell module \citep{Stensola2012}. Multiple computational models of grid cells exist that show how they might perform path integration or dead reckoning \citep{Giocomo2011}. This model does not depend on which of these implementations grid cells are using, it only depends on grid cells being able to update given movement information.

\textbf{Grid cells that represent location of object in environment:} This model uses a second population of grid-like cells that tracks locations of objects rather than self-location. Following basic intuition of mapping systems, it is theoretically likely that objects are associated with their locations on the grid, but it is an open question whether any grid-like cells would need to be dedicated to representing this location. A set of experiments has shown a population of grid cells that track viewed location \citep{Killian2012, Nau2018, Julian2018a} and covertly attended locations \citep{Wilming2018}. One possible interpretation of these empirical results is that the recorded grid cells were in fact these theorized grid cells. In this model, this location acts as a routing signal between object-vector cells and grid cells, and it is computed by detecting which object-vector cells fire with which grid cells.

\textbf{Displacement cells:} Based on theoretical considerations, it has been proposed that there are likely neurons that detect the spatial displacement between two metric location representations \citep{Bush2015, Hawkins2019}. This model uses the same mechanism as previously proposed, with a population of displacement cells responding to or invoking sudden changes in a metric location representation. One noteworthy difference in this model is that the displacement cells detect sudden changes in object-vector cells, rather than sudden changes in grid cells.

\textbf{Memory nodes:} Index theory of hippocampus suggests that some subset of hippocampal neurons form representations whose job is simply to learn associations with multiple other populations \citep{Teyler1986, Josselyn2020}. The representations themselves act as an index (similar to an index of an array) for recalling all attributes of a memory. Neurons in these representations, sometimes referred to as engram cells, are often hypothesized to carry no meaning of their own. Rather, they are given meaning via learned associations to populations that carry the content of the memory. For example, in this model these cells represent instances of environment parts. If an agent sees an environment part in Environment 1, it will create a memory node for this instance of this environment part, and it associates with it a description of the part, and it also associates its location and its displacement relative to other nearby parts. This content is all stored by learning associations with memory node representations. If a similar environment part is encountered in Environment 2, a different memory node will be allocated to that part and its attributes.

\subsubsection{How this model combines these building blocks}
\label{section:combined-building-blocks}

\begin{figure}[]
\begin{center}
\includegraphics[width=0.9\textwidth]{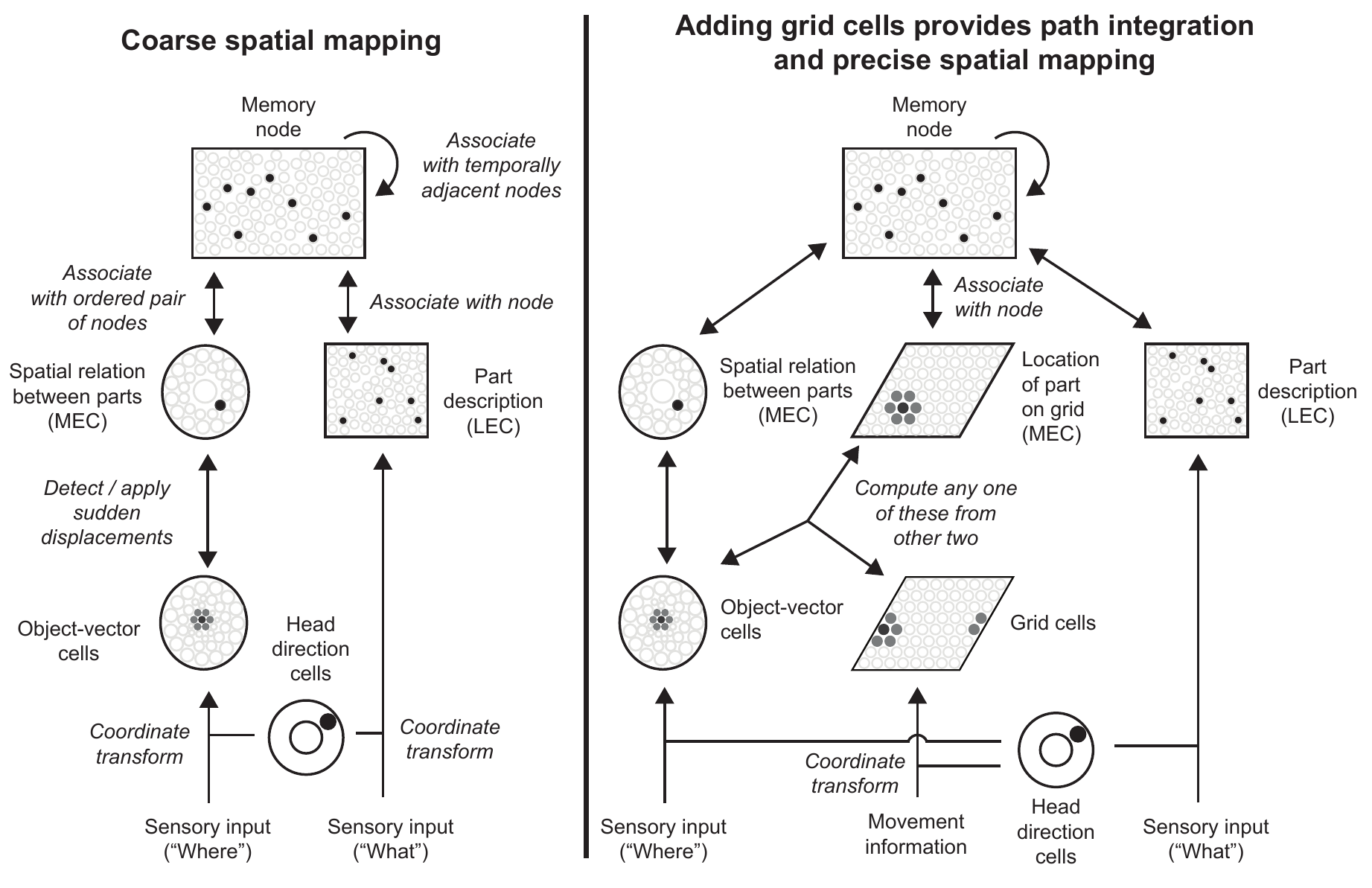}
\end{center}
\caption{\label{fig:circuit}\textbf{Neural mechanism for quickly building memory graphs.} Spatial mapping can work without grid cells. Adding grid cells makes the spatial map more precise and enables path integration (also known as ``dead reckoning''). \textit{(Left)} A minimal circuit. Information from sensory input goes through a coordinate transformation, resulting in object-vector cells and an allocentric description of the sensed part. Spatial relations between object-vector cells are simply coarse displacement vectors. Graph nodes are represented by random sparse hippocampal activity patterns. Graph nodes are associated with other temporally adjacent graph nodes, enabling the system to either oscillate between nodes or allow the attention system to know which nodes can be attended to. After relation information has been learned, object-vector cells can be coarsely predicted during attention shifts, but they cannot be precisely predicted. \textit{(Right)} The same model, now with grid cells added. Movement information goes through a similar coordinate transformation and is used to update grid cells. Using grid cells and object-vector cells, the location of the sensed part on the grid is detected and associated with the graph node. The coarse relation between parts and the grid location of each part can together be used to accurately predict object-vector cells during movement and during attention shifts.}
\end{figure}

This model can be understood by first considering what it can do without grid cells, then adding grid cells \textbf{(Figure \ref{fig:circuit})}.

In this model, self-location is primarily represented by object-vector cells in conjunction with a graph node. The network represents one environment part at a time, quickly switching between representing different environment parts. This switching can be interpreted as attention shifts or as fast oscillations between nodes, enabled by learned associations between temporally adjacent nodes. Each time a node becomes active, it drives the part description to become active. Additionally, each ordered pair of nodes drives a learned relation between parts to become active. During attention shifts, these learned relations can be used to coarsely predict which object-vector cell should become active.

Now we add grid cells. Similar to before, self-location is primarily represented coarsely by object-vector cells and a graph node, but grid cells provide a complementary uniform resolution location, augmenting the agent's location representation with extra precision. In this extended model, the grid location of each environment part is associated with graph nodes. This information enables the model to more precisely predict object-vector cells during attention shifts. Grid cells receive movement information and perform path integration, and these updates are propagated to the object-vector cells.

\begin{figure}[]
\begin{center}
\includegraphics[width=0.9\textwidth]{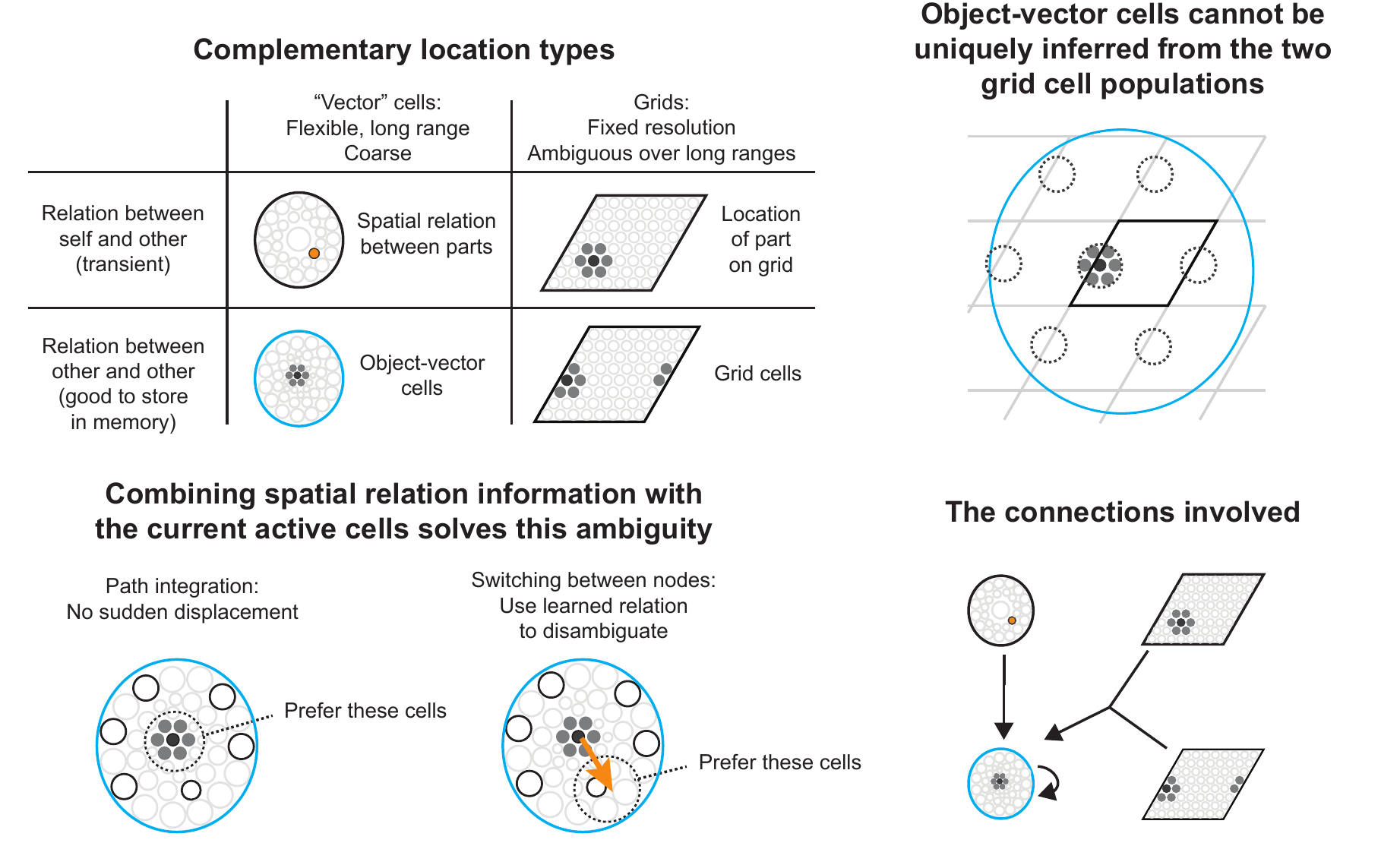}
\end{center}
\caption{\label{fig:ambiguity}\textbf{Grid cells' ambiguity complements the ambiguity of relations.} \textit{(Top-left)} A 2x2 taxonomy of this model's representations of location. Vector-like cells use varying resolution, making them long-range but coarse. Grid-like representations are ambiguous over long ranges, but they use uniform resolution. \textit{(Top-right)} Depiction of an example ambiguous scenario. The blue circle indicates how the object-vector cells are mapped onto space, while the black rhombus indicates the grid cells, which tile the space. The \textit{Location of part on grid} describes where the center of the blue circle lands on the black rhombus, giving us this picture. As shown here, the grid cells alone do not uniquely determine which object-vector cells should be active. Different sets of object-vector cells that are consistent with the active grid cells are shown as dotted circles. \textit{(Bottom-left)} This problem is solved by using the memory graph's coarse displacement information. During path integration, the object-vector cells give preference to cells that are near the previously active cells. When switching between nodes on the memory graph, the orange coarse displacement information is used to determine which subset of the object-vector cells should be given preference. Thus, the \textit{Location of part on grid} and \textit{Coarse displacement} work together to enable mapping spaces much larger than one grid cell period and potentially much larger than the object-vector cell range. \textit{(Bottom-right)} This operation of predicting object-vector cells requires all of the other populations, and the computation also depends on the current active object-vector cells, as indicated by the recurrent arrow.}
\end{figure}

An important design principle of this model is that it uses different spatial resolutions in grid-like vs.\@ vector-like populations \textbf{(Figure \ref{fig:ambiguity})}. In this model, grid cells use a fixed spatial resolution so that they can perform path integration, whereas object-vector cells use varying resolution (which is speculative -- not empirically verified) so that they can be distributed over a larger area using a relatively small number of cells. This division of labor enables the object-vector cells to be path-integrated by the grid cells, and it allows precise coding of locations that are far from environment parts via the combined representations of the coarse-but-unambiguous object-vector cells and the precise-but-ambiguous grid cells. Similarly, displacement vectors between objects are stored using varying resolution while the grid locations of objects are stored in fixed resolution, and so the two combined learn the precise relative locations of objects. Considering the fundamentals of sensory processing, this design is a natural fit in multiple ways. First, as a viewer moves away from an object, the image of the object on the retina changes more slowly, so it is natural for sensory-driven cells like object-vector cells to have larger and larger fields as distance increases. Second, displacement vectors between objects can be easily identified coarsely, but often not precisely, so it is pragmatic for the system to have one set of cells that quickly learn the coarse displacements between objects, while another set of cells learn the precise relative locations more slowly by learning the objects' grid locations, similar to how robots iteratively update locations of objects when performing Simultaneous Localization and Mapping (SLAM).

A core capability of this model is its ability to update object-vector cells during movement and during attention shifts. These operations can be performed precisely when the model uses grid cells. This relies on being able to route between grid cells and object-vector cells given the grid location of an object, shown as a three-way arrow in \textbf{Figure \ref{fig:circuit}}. This is built on the intuitive property that if an agent knows its location on the grid and it also knows an object's distance and direction from the agent, it can infer the object's location on the grid. The required connectivity for this three-way circuit is similar to that of cells that detect and apply displacements to grid cells \citep{Bush2015, Hawkins2019} or rotations to vector cells \citep{Bicanski2018}. Each cell receives projections from pairs of cells in the other populations so that they receive strong excitatory input when that pair of cells becomes active. However, in this model there as an additional complication, due to the ambiguity of grid cells \textbf{(Figure \ref{fig:ambiguity})}. Because this model's object-vector cells can cover a much larger area than one grid cell spatial period, it is not possible for this three-way circuit to unambiguously infer object-vector cells from grid cells. This problem is solved by making the object-vector cells' activations competitive and using learned coarse displacements to supply them with additional information. During path integration, the grid cells update in response to movement, and this update is propagated to the object-vector cells. Because no node switches occur, no displacement cells are activated, so object-vector cells that represent locations near that of previously active cells are given preference, solving the ambiguity problem during path integration. This leaves the problem of ambiguity when rapidly switching between objects. This rapid switch causes displacement cells to activate, providing excitatory input to a certain subset of the object-vector cells, depending on which ones are currently active. This excitatory input combined with the excitatory input from the three-way circuit will activate the appropriate object-vector cell. This mechanism demonstrates how object-vector cells can be updated by grid cells, and it demonstrates how object-vector cells can rapidly switch between representing different objects if the system has learned those objects' grid locations and the coarse displacements between those objects.

Putting these pieces together, we get a human-interpretable mapping system that readers will be able to mentally simulate. Consider the scenario of an animal arriving in a novel environment. The agent already has a grid cell and head direction cell representation active (or, if not, it can randomly activate one). The agent sees an object, inferring some description of the object and the agent's location relative to that object, and it uses head-direction cells to perform a coordinate transformation, inferring allocentric object-vector cells. Using the current active grid cells and object-vector cells, the agent infers the object's location on the grid. Because the location of the object on the grid is now known, updates to grid cells will drive updates to object-vector cells without the agent needing to sense the object again, enabling prediction during self-motion. We continue our mental simulation, now focusing on how the map is learned. The agent forms a new memory, allocating a memory node and associating it with this object's description and its location on the grid. The agent now senses a second object in the environment, going through the same above process. This drives a different object-vector cell to suddenly become active, driven by this sensory input. The displacement between the vectors (i.e.\@ between this object-vector cell and the previously active object-vector cell) is detected by a set of displacement cells, and this displacement is associated with this pair of memory nodes. Finally, the agent attends to the first object. The map ought to enable the agent to anticipate which object-vector cells should become active. Using the combination of the currently active object-vector cells, the learned coarse displacement information, the currently active grid cells, and the learned location of the object on the grid, the correct newly active object-vector cells can be anticipated. This can feed backward into the sensory system, enabling prediction. Thus, as the agent attends to different objects, grid cells and head direction cells remain stable while all the other populations switch between objects. Continuing this learning process, the system can learn arbitrarily large maps as long as the environment-parts are not long distances apart (i.e. longer than the object-vector cell range).

\subsection{Graph processing}
\label{section:graph-processing}

By moving from a system built on Cartesian coordinates analogs to a system built on graphs, we gain generality, but we also lose some conveniences of Cartesian coordinates. With Cartesian coordinates, after each environment part is attended to once, the system can trivially compute displacements between any pair of parts. In a graph-based system, computing a displacement between two parts either requires having previously observed those parts in that order, or it requires finding a path between them in the graph and aggregating over that path. We even see this issue coming up in the mental simulation from the previous paragraph. In that example, when the agent attended back to the first object, it had never previously experienced the ordered transition from Object 2 to Object 1, it had only experienced Object 1 to Object 2. It is trivial on a computer to infer this reversed displacement, but can the brain's neural circuitry do it?

It an open interesting question what types of computations the brain performs on these graphs. There are three broad possibilities. In one extreme, the system may simply require all relevant transitions to be learned by brute force. If, in practice, transitions tend to happen in certain orders, this may be sufficient. In the other extreme, the system may be able to learn a minimal graph and perform advanced computations on it during inference. This may involve a form of replay \citep{Foster2017}, with different readout neurons performing different computations on the replayed sequence. Finally, in a middle-ground possibility, the brain could quickly learn a minimal graph, then later during rest or sleep fill in missing edges, possibly relying on replay.

\section{Predictions of the model}

In this paper I have argued that the hippocampal formation learns environments as arrangements of familiar parts, and in particular that these arrangements of parts are learned as relation graphs. One high-level implication of this idea is the hippocampal formation does not need to form an association at each location in an environment. In the model of \cite{Whittington2020}, every place cell population code is learned as an attractor state, so every place cell representation that occurs in the environment is memorized. In this model, only the graph of environment parts is memorized, so learning the environment does not involve learning these place cell attractor states. Depending on how ``place cells'' are defined, it is possible that the memory index / engram cells representing nodes in this graph will be classified as place cells. These cells do not in fact encode self-location, rather they become active when the agent is attending to a particular environment part, though this may correlate strongly with certain locations.

Diving deeper into the particulars of this model, further predictions can be made. This model suggests that grid cells tracking eye position \citep{Killian2012, Nau2018, Julian2018a} and covertly attended locations \citep{Wilming2018} are actually tracking locations of attended objects, or the locations of the subject of any other type of ``vector'' cells \citep{Bicanski2020}, e.g.\@ goals. Object-vector cells have been seen to respond at multiple locations in environments that have multiple objects \citep{Hoydal2019}, and this model suggests that object-vector cells will either respond to different objects at different times in the theta band, quickly oscillating between multiple objects, or will shift with attention -- either way, there will be time-based separation. It is important to note that this paper focused on interpretable models and toy datasets. Actual neural tuning curves will be more complicated.

\section{Discussion}

The hippocampal formation is often described as a spatial mapping system. It is worth asking whether spatial mapping is the base algorithm of the system, or if there is some other base algorithm and spatial mapping is just one of its applications. With this model, I aimed to frame spatial mapping as just one instantiation of a more general fast-learned-relation-graph algorithm. By allowing the repertoire of relations to be learned, this system becomes highly general, applicable to spatial and non-spatial problem domains. In this paper I did the preliminary work of framing spatial mapping as a graph-learning process. Future work will exploit this gained generality.

This point of view is compatible with the high-level point of view of TEM \citep{Whittington2020}, but it differs in some important details. Using the language of this paper, TEM attempts to capture all relations by associating location information with graph nodes, not by associating information with graph edges. I argue that, though node-only associations are desirable when possible, it will not always be possible, and the ability to associate arbitrary relation information with graph edges is important. It is likely that a slight variation on TEM could learn exactly the model that I have presented here.

\section*{Acknowledgements}

Many thanks to Mirko Klukas, Jeff Hawkins, and Subutai Ahmad for helpful discussions.

\bibliographystyle{apa}
\bibliography{references}

\end{document}